\documentclass[letterpaper, 10 pt, conference]{ieeeconf}
\IEEEoverridecommandlockouts
\usepackage{cite}

\usepackage{graphicx}
\usepackage{multirow}
\usepackage{dblfloatfix}

\usepackage[bookmarks=true]{hyperref}
\hypersetup{%
  breaklinks,
  bookmarksnumbered=true,
  bookmarksopen=true,
  colorlinks=true,
  linkcolor=black,
  urlcolor=black,
  citecolor=black
}

\usepackage{amsmath}
\usepackage{amsfonts}
\usepackage{comment}
\usepackage{algorithm}
\usepackage[noend]{algpseudocode}

\algdef{SE}[DOWHILE]{Do}{doWhile}{\algorithmicdo}[1]{\algorithmicwhile\ #1}%

\algrenewcommand\algorithmicindent{0.8em}

\algdef{SE}[SUBALG]{Indent}{EndIndent}{}{\algorithmicend\ }%
\algtext*{Indent}
\algtext*{EndIndent}

\usepackage{array}

\makeatletter
\newcommand{\thickhline}{%
    \noalign {\ifnum 0=`}\fi \hrule height 1.5pt
    \futurelet \reserved@a \@xhline
}
\newcolumntype{"}{@{\hskip\tabcolsep\vrule width 1.5pt\hskip\tabcolsep}}
\makeatother

\begin{document}
\bstctlcite{IEEEexample:BSTcontrol}

\title{\LARGE \bf Convergent iLQR for Safe Trajectory Planning and Control \\ of Legged Robots}

\author{James Zhu, J. Joe Payne, and Aaron M. Johnson %
    \thanks{This material is based upon work supported by the National Science Foundation
    under grant \#CMMI-1943900.}%
    \thanks{All authors are with the Department of Mechanical Engineering, Carnegie Mellon University, Pittsburgh, PA, USA, \texttt{jameszhu@andrew.cmu.edu}, \texttt{amj1@andrew.cmu.edu}}%
}

\maketitle
\thispagestyle{empty}
\pagestyle{empty}


\begin{abstract}
    In order to perform highly dynamic and agile maneuvers, legged robots typically spend time in underactuated domains (e.g.\ with feet off the ground) where the system has limited command of its acceleration and a constrained amount of time before transitioning to a new domain (e.g.\ foot touchdown).
    Meanwhile, these transitions can instantaneously change the system's state, possibly causing perturbations to be mapped arbitrarily far away from the target trajectory.
    These properties make it difficult for local feedback controllers to effectively recover from disturbances as the system evolves through underactuated domains and hybrid impact events.
    To address this, we utilize the fundamental solution matrix that characterizes the evolution of perturbations through a hybrid trajectory and its 2-norm, which represents the worst-case growth of perturbations.
    In this paper, the worst-case perturbation analysis is used to explicitly reason about the tracking performance of a hybrid trajectory and is incorporated in an iLQR framework to optimize a trajectory while taking into account the closed-loop convergence of the trajectory under an LQR tracking controller.
    The generated convergent trajectories recover more effectively from perturbations, are more robust to large disturbances, and use less feedback control effort than trajectories generated with traditional  methods.
\end{abstract}


\section{Introduction}
        
    Legged robotics research has increasingly focused on enabling highly dynamic and agile motions such as jumping, leaping, and landing \cite{paper:johnson-icra-2013,kolvenbach_moon_jumping,nguyen_jumping_2021,li_jumping_2023}.
    Implementing these capabilities reliably would improve legged robot performance in applications such as extraterrestrial or urban environment navigation where jumping up on ledges or leaping across chasms may be necessary.
    However, jumping and leaping are dangerous maneuvers, with failure often resulting in catastrophic outcomes for the robot.

    What makes these actions challenging is that they induce trajectories that are both hybrid and underactuated, which doubly contribute to the difficulty in controlling legged robots.
    Broadly speaking, a system is hybrid if it undergoes discrete changes in state and/or dynamics \cite{vanderschaft_hybrid_systems,lygeros_hybrid_automata}, and it is underactuated if there exists a direction of acceleration in state space that can not be commanded by any valid input \cite[Ch. 1.2]{underactuated}.    
    Even when an underactuated system is controllable, driving the system to a desired target state may require significant time and control effort, neither of which may be readily available.
    For instance, a robot jumping in the air can not arbitrarily choose how much time it has until its feet touchdown on the ground.
    This means that the controller needs to spend a lot of effort to correct tracking errors prior to touchdown, or else discontinuous, unbounded saltation effects \cite{kong2023saltation} can cause arbitrarily large divergence if incoming errors are not sufficiently mitigated, e.g.\ with grazing impacts.
    Increasing control gains is one possible solution to improve stability, though that strategy comes at a large drawback of worsening robustness in the face of modelling errors and uncertainties\cite[Ch. 13]{Haidekker2020}.
    Instead, this work leverages nonlinearities in continuous and hybrid dynamics that make some trajectories easier to stabilize than others, even under equivalent feedback controllers.
    
    \begin{figure}[t]
        \centering
        \vspace{-.5em}
        \includegraphics[width=1\linewidth]{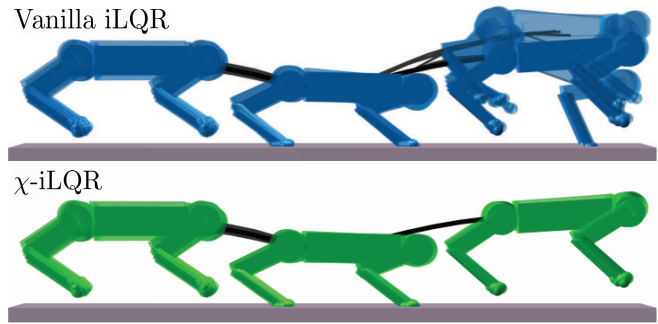}
        \vspace{-1.5em}
        \caption{Similar quadruped gaits tracked with equivalent LQR controllers display enormous differences in final tracking performance. Results for 50 paired trials of trajectories sampled from an initial error covariance of $10^{-2}$ in all directions are shown. The top blue trajectory was generated with a standard (vanilla) iLQR algorithm, while the bottom green trajectory was generated with our novel $\chi$-iLQR, which improves the average closed-loop convergence by 92\%. The horizontal distance between the displayed frames is exaggerated for visual clarity. Only the convergence of the position states is represented, and does not indicate convergence of the velocity states.}
        \label{fig:quad_comparison}
    \end{figure}

    This paper presents a novel adaptation of the iLQR trajectory optimization algorithm that improves closed-loop convergence under an equivalent feedback controller (i.e.\ without changing the LQR controller weights), as demonstrated in Fig. \ref{fig:quad_comparison}. 
    Our simulation results show that this convergent iLQR ($\chi$-iLQR) achieves three simultaneous improvements over standard iLQR: superior tracking performance from initial perturbations, reduced feedback control effort over the trajectory, and improved robustness to large initial errors.
    Compared to existing methods, $\chi$-iLQR has two additional key strengths.
    Firstly, it is based on an analysis that is simple to compute compared to methods such as sum-of-squares.
    Additionally, $\chi$-iLQR captures the local tracking performance of a closed-loop trajectory, which directly predicts experimental results.

\section{Related Works}
    A strategy that has been used to enable dynamic yet precarious behaviors for legged robots is leveraging highly accurate, complex models and full-body trajectory optimization to plan precise motions \cite{nguyen_jumping_2021,norby_adaptive_complexity_2022}.
    While these methods incorporate feedback controllers to stabilize the generated trajectories, there has been little focus on how these feedback controllers should be designed to stabilize closed-loop systems under error and uncertainty.

    Robust trajectory planning has been successfully implemented for smooth systems like wheeled robots, with some recent results being adapted to hybrid systems like legged robots.
    These works have focused on optimizing over uncertainties in system dynamics, such as unknown disturbances and modelling errors \cite{manchester_ditrel_2019,morimoto_minimax_ddp_2003,lew_chance_constrained_2020}.
    For example, \cite{manchester_ditrel_2019} designs robust closed-loop trajectories for smooth systems by optimizing the volume reduction of an ellipsoidal disturbance set, but was not applied to hybrid systems.
    Risk-sensitive planning and control is an alternate method that optimizes over the variance of a cost distribution that evolves through the trajectory \cite{hammoud_risk_sensitive_2021, ahmadi2020risk}.
    Other approaches present trajectory optimization algorithms for legged robots over uncertain terrain \cite{drnach_robust_2021} and compute a forward reachable set to bound closed-loop errors \cite{kousik_safety_2020}.
    Many of these methods require the distribution of errors to be prespecified, which is not always clear how to tune.
    Additional actuation, such as reaction wheels or tails, also relieves the difficulties of underactuated systems \cite{lee_balance_2023, yang2023proprioception}.
    However, this comes with obvious tradeoffs of increased cost, size, and weight.

    Separately, consider the problem of quantifying the stability or convergence properties of a system.
    A very popular method is Lyapunov analysis, where the existence of a positive definite differentiable scalar function with negative definite derivatives, called the Lyapunov function, can guarantee asymptotic stability of the system \cite{Khalil}.
    Lyapunov functions can be difficult to compute, particularly for hybrid systems, and can require methods such as sum-of-squares \cite{papa_sum_of_squares_2002} or machine learning \cite{chen_learning_lyapunov} to be tractable.
    A similar strategy known as control barrier functions, which restricts the system from entering some set of undesirable states, has been successfully implemented on legged robot hardware \cite{grandia2021multi}, but has the same drawback as Lyapunov functions.
    
    A different strategy to analyze the stability and convergence of trajectories is contraction analysis \cite{contraction_analysis}, which tracks the distance between two close trajectories.
    If this distance monotonically decreases over the trajectory, then the system is contractive and asymptotic stability can be guaranteed \cite{contraction_analysis}.
    Contraction analysis has been incorporated into path planning and trajectory optimization algorithms on smooth systems \cite{johnson2016convergent, kong2019optimally}, but applying contraction analysis to hybrid systems is difficult because many mechanical hybrid systems are not contractive at hybrid events \cite{burden_contraction}.
    \cite{zhu_hybrid_event_shaping} loosened the contraction criterion and  optimized the stability of open-loop periodic orbits using monodromy matrix analysis.
    Here, we extend that work by generalizing to non-periodic trajectories under feedback control.

\section{Analysis and Planning for Hybrid Systems}


This section defines a hybrid system and quantifies the performance of a closed-loop hybrid trajectory using linearized variational equations.
With this analysis, we can generate a scalar measure of a trajectory's convergence which is then incorporated into a trajectory optimization framework.
    \subsection{Hybrid Systems}
        
    Hybrid systems are a class of dynamical systems that consist of continuous domains connected by hybrid events\cite{vanderschaft_hybrid_systems,lygeros_hybrid_automata}. 
    Following the notation in \cite{kong2023saltation}, we describe a hybrid system as a set of discrete modes $\{\mathrm{I},\mathrm{J},\hdots,\mathrm{K}\}$, each with a domain ${D}_\mathrm{I}$
    and a time-varying vector field  $F_\mathrm{I}$. $G_{(\mathrm{I},\mathrm{J})}$ is a guard that triggers a transition between mode $\mathrm{I}$ and mode $\mathrm{J}$ and $R_{(\mathrm{I},\mathrm{J})}$ is the reset map defining that transition.
    
    An execution of a hybrid system \cite{johnson_hybrid_systems} begins at an initial state $x_0\in D_\mathrm{I}$. 
    With input $u_\mathrm{I}(t,x)$, the system obeys the dynamics $F_\mathrm{I}$ on $D_\mathrm{I}$. 
    If the system reaches guard surface $G_{(\mathrm{I},\mathrm{J})}$, the reset map $R_{(\mathrm{I},\mathrm{J})}$ is applied and the system continues in domain $D_\mathrm{J}$ under the corresponding dynamics defined by $F_\mathrm{J}$. 
    The flow $\phi(t,t_0,x_0,U)$ describes how the hybrid system evolves from some initial time $t_0$ and state $x_0$ until some final time $t$ under input sequence $U$.

    \subsection{Linearized Variational Equations}
     \label{sec:lin_var}
     For both continuous domains and hybrid transitions, linearized variational equations can be constructed to characterize the evolution of perturbations $\delta x$ \cite{hirsch2012differential}.
     In each continuous domain, the linearized variational equation is discretized from timestep $i$ to $i+1$ and is $\delta x_{i+1}\approx (A_\mathrm{I}-B_\mathrm{I}K_\mathrm{I})\delta x_i$ \cite{hirsch2012differential} with $A_I$ and $B_I$ being the derivatives of the discretized dynamics in mode $\mathrm{I}$ w.r.t.\ state $x_i$ and control inputs $u_i$, respectively, and $K_\mathrm{I}$ are linear feedback gains.
     The feedback term drops out for open-loop systems.
     For hybrid events, the analogous variational equation is the saltation matrix $\Xi_{(\mathrm{I},\mathrm{J})}$, which describes the transition between modes $\mathrm{I}$ and $\mathrm{J}$.
     The saltation matrix is the first-order approximation of the change in state perturbations from before the hybrid event at $\delta x(t^-)$ to perturbations after $\delta x(t^+)$ \cite{burden_contraction}, such that $\delta x(t^+) \approx \Xi_{(\mathrm{I},\mathrm{J})}\delta x(t^-)$.
     This linear approximation assumes that nearby trajectories undergo the same mode transition.
    Computing the saltation matrix relies on the derivatives of the reset and guard of the transition along with the dynamics in each mode, and is detailed in \cite{kong2023saltation}.

    \subsection{Convergence Measure}
    Consider a trajectory that begins at state $x_0 = x(t_0)$ at time $t_0$ and is executed until time $t_f$ where it arrives at state $x_f = \phi(t_f,t_0,x_0,U)$.
    The control objective is to bring any nearby initial state $\bar{x}_0 = x_0 + \delta x_0$ towards the nominal trajectory so that at time $t_f$, $\bar{x}_f = \phi(t_f,t_0,\bar{x}_0,\bar{U}) = x_f + \delta x_f$ is closer to $x_f$.
    To characterize the closeness of $\bar{x}_f$ and $x_f$, we utilize the fundamental solution matrix, $\Phi$. Following \cite{leine_dynamics}, the fundamental solution matrix is the linearized approximation
    $\delta x_f \approx \Phi\delta x_0$ and represents the transformation of error from the initial state to final state.
    
    The fundamental solution matrix can be computed by sequentially composing the linearized variational equations in each continuous domain ($\tilde{A} := A-BK$) and the saltation matrices ($\Xi$) at each hybrid event \cite{zhu_hybrid_event_shaping}. 
    For a hybrid trajectory with $N$ domains, the fundamental solution matrix is:
    \begin{align}
        \Phi &= \tilde{A}_N\Xi_{(N-1,N)}\dots \Xi_{(2,3)}\tilde{A}_2\Xi_{(1,2)}\tilde{A}_1
        \label{eq:phi}
    \end{align}
    
    Since the fundamental solution matrix captures the change in errors across a trajectory, the singular values of $\Phi$ characterize error change along principle axes of state space. 
    The largest singular value, which is equivalent to the induced 2-norm of $\Phi$, describes the evolution of the most divergent direction of initial error $\delta x_0$.
    We define the convergence measure, $\chi$ to be exactly this worst-case value:
    \begin{align}
        \chi = ||\Phi||_2
    \end{align}
    
    $\chi$ is a continuous measure of local convergence, where smaller values of $\chi$ indicate stronger reduction of worst-case final errors.
    A value of $\chi < 1$ indicates errors in all directions will shrink.
    
    \subsection{iLQR for Hybrid Systems} 
    The iterative linear quadratic regulator (iLQR) is a trajectory optimization method that also computes LQR feedback gains over the generated trajectory \cite{todorov_ilqr}.
    iLQR is convenient because compared to other trajectory optimization methods like direct collocation \cite{kelly_optimization}, it is less computationally intensive and guarantees a feasible trajectory.
    We draw from recent work that adapts the iLQR algorithm for use on hybrid dynamical systems \cite{kong2021ilqr,paper:kong-hybrid-2023}.
    
    In brief, iLQR solves the optimal control problem over $N$ discretized timesteps:
    \begin{align} 
        \min_{U} \quad & \ell_N (x_N) + \sum_{i=0}^{N-1} \ell_i(x_i, u_i) \\ 
        \text{where} \quad & x_0 = x(0)\\
        & x_{i+1} = \phi(t_{i+1},t_i,x_i,u_i) & \forall i
        \label{eq:dynamics_constraint}
    \end{align}
    where $\ell_i(x_i,u_i)$ and $\ell_N(x_N)$ represent the nonlinear stage cost and terminal cost, respectively, $X := \{x_0,x_1,...,x_{N}\}$ is a sequence of states with $x_i \in \mathbb{R}^n$ the system state at timestep $i$ and $U := \{u_0,u_1,...,u_{N-1}\}$ is a sequence of control inputs with $u_i \in \mathbb{R}^m$ the control input at timestep $i$.
    We also record the sequence of domains $M := \{D_0,D_1,...,D_{N}\}$ with $D_i$ the domain at timestep $i$ such that $x_i \in D_i$.
    $\phi$ is the aforementioned flow of the trajectory.
    
    iLQR computes gradient and Hessian information of the cost, which results in a quadratic approximation of the cost function.
    As such, the state and terminal costs can equivalently be simplified as quadratic functions
    such that the cost function is simplified to:
   \begin{align}
        J = x_N^TQ_Nx_N + \sum_{i=0}^{N-1} x_i^TQ_ix_i + u_i^TR_iu_i
        \label{eq:vanilla_cost}
    \end{align}
    with $Q_i,Q_N \in \mathbb{R}^{n\times n}$ and $R_i \in \mathbb{R}^{m\times m}$ all positive definite.
    
    iLQR solves the optimal control problem by alternating between forward passes that simulate the system under a given control input sequence, and backward passes that solve for a new locally optimal control sequence.
    In the backward pass, the value function, which is the optimal cost to go at any timestep, is propagated through the trajectory in reverse, and gives locally optimal feedforward inputs and feedback gains at each timestep. 
    Computing the value function relies on gradient and Hessian computations of the cost function and Jacobians of the dynamics, which equates to computing the linearized variational equations discussed in Sec. \ref{sec:lin_var}.
    For much greater detail of iLQR for hybrid systems, see \cite{kong2021ilqr,paper:kong-hybrid-2023}.
    
    \section{Convergent iLQR}
    
    Here we present a novel trajectory optimization algorithm called convergent iLQR or $\chi$-iLQR, summarized in Algorithm \ref{alg:convergent}. 
    In this method, the convergence measure is added to the cost function from \eqref{eq:vanilla_cost} such that the algorithm minimizes:
    \begin{align}
        J_{\chi} = \quad & Q_{\chi}\chi + x_N^TQ_Nx_N + \sum_{i=0}^{N-1} x_i^TQ_ix_i + u_i^TR_iu_i
        \label{eq:convergent_cost}
    \end{align}
    where $Q_\chi$ is a scalar weighting parameter.
    Since $\chi$ is solely a function of states and inputs, iLQR uses gradient and Hessian information to make a quadratic approximation compatible with the other cost terms.
    
    Typically in iLQR, the cost function $J$ is evaluated after each forward pass, since it is only dependent on the states and inputs of the most recent trajectory.
    However, in this case the convergence measure portion of the cost function is dependent on the feedback gains generated by the algorithm.
    This means that the gradient and Hessian terms of the cost function rely on the feedback gains that are being updated at every timestep in the backward pass.
    Due to this, the cost function derivatives are highly coupled with the gains and become convoluted to compute.
    
    To resolve this, we propose executing two separate backward passes that each compute a different set of gains.
    We do this to preserve the convergence properties of iLQR, though other choices might be possible such as borrowing the previous set of gains under the assumption that the gains do not change significantly over iterations.
    First, the tracking backward pass computes the feedback gains that will be used as the LQR tracking controller gains and to compute the convergence measure. It is equivalent to the backward pass in standard iLQR using the cost function $J$ \eqref{eq:vanilla_cost}, which solves the Riccati equation for the most recent trajectory.
    With the gains generated in the tracking backward pass $K_t$, the convergent cost function $J_\chi$ \eqref{eq:convergent_cost} can be computed.
    The search backward pass takes $J_\chi$ from the tracking backward pass and computes the gradients of the convergent cost function with controller gains $K_t$.
    The feedforward inputs $k_s$ and the feedback gains $K_s$ from this pass are used to search for an improved trajectory in the forward pass. 
    
    Since $J_\chi$ is returned by the tracking backward pass, a line search is performed after this function call to guarantee the reduction of the cost function $J_\chi$.
    If the line search condition is not satisfied, the forward pass and tracking backward pass are looped until the line search condition is passed.
     
    Within the search backward pass, iLQR requires computation of the gradient and Hessian of $\chi$.
    The derivatives of $\chi$ can be computed by leveraging the singular value decomposition of $\Phi = USV^T$ where $S$ is a diagonal matrix of singular values and the columns of $U$ and $V$ are the left and right singular vectors, respectively.
    $\chi$ is the largest singular value of $\Phi$ and let $u_\chi$ and $v_\chi$ be its corresponding left and right singular vectors.
    Following \cite{townsend_singular_diff}, the derivative of $\chi$ with respect to the state at timestep $i$ is:
    \begin{align}
        \frac{\partial\chi}{\partial x_i} &= u_\chi^T\frac{\partial\Phi}{\partial x_i}v_\chi
    \end{align}
    $\frac{\partial\Phi}{\partial x_i}$ in turn can be computed by using the product rule along with leveraging the fact that only $\tilde{A}_i$ and $\Xi_{(i,i+1)}$ are functions of $x_i$, and all other $\tilde{A}$ and $\Xi$ terms have zero derivatives with respect to $x_i$.
    For notational brevity, let:
    \begin{align}
        P_{i} &= \tilde{A}_N\cdots\Xi_{(i+1,i+2)}\tilde{A}_{i+1} \\
        O_{i} &= \Xi_{(i-1,i)}\tilde{A}_{i-1}\cdots\Xi_{(1,2)}\tilde{A}_1
    \end{align}
    such that $\Phi = P_{i}\Xi_{(i,i+1)}\tilde{A}_iO_{i}$
    and
    $\frac{\partial P_{i}}{\partial x_i} = \frac{\partial O_{i}}{\partial x_i} = 0$.
    Thus:
    \begin{align}
        \frac{\partial\Phi}{\partial x_i} &= P_{i}\frac{\partial\Xi_{(i,i+1)}}{\partial x_i}\tilde{A}_iO_{i} + P_{i}\Xi_{(i,i+1)}\frac{\partial\tilde{A}_i}{\partial x_i}O_{i}
        \label{eq:product_rule}
    \end{align}
    Derivatives with respect to the input $u_i$ follow equivalently. To improve computational efficiency, $O_{i}$ at each timestep can be computed recursively during the forward pass and each $P_{i}$ can be computed recursively in the tracking backward pass.
    Since the rollout does not yet have feedback gain information, the initial $O$ values must be computed separately.

    \begin{algorithm}[t]
        \caption{Convergent iLQR Algorithm}
        \label{alg:convergent}
        \begin{algorithmic}
            \State Initialize $U$, $Q_\chi$, $Q_N$, $Q_i$, $R_i$, $n_{\text{iterations}}$
            \State $X$, $U$, $M$, $J \gets$ \Call{Rollout}{$U$}
            \State $K_t$, $J_\chi$, $P \gets$ \Call{TrackingBP}{$X$, $U$, $M$, $J$}
            \State $O \gets$ 
            \Call{ComputeO}{$X$, $U$, $M$, $K_t$}
            \For{$i \gets  1 \text{ to } n_{\text{iterations}}$}
                \State $k_s$, $K_s \gets$ \Call{SearchBP}{$X$, $U$, $M$, $J_\chi$, $O$}
                \Repeat
                    \State $X$, $U$, $M$, $J$, $O \gets$ \Call{ForwardPass}{$X$, $U$, $M$, $k_s$, $K_s$}
                    \State $K_t$, $J_\chi$, $P \gets$ \Call{TrackingBP}{$X$, $U$, $M$, $J$}
                \Until \Call{LineSearchIsSatisfied}{$J_\chi$}
            \EndFor
            \Return $X$, $U$, $M$, $K_t$
        \end{algorithmic}
    \end{algorithm}
    
    Because the scalar $\chi$ is derived from the norm of the matrix $\Phi$, the gradient of $\chi$ relies on computing a 3-dimensional tensor of $\tilde{A}_i$ and $\Xi$ derivatives, and the Hessian of $\chi$ is computed from a 4-dimensional tensor of matrix second derivatives.
    While recent work has enabled faster computation of second derivatives of dynamics \cite{singh_ddp_derivatives_2023} which can aid in the computation of the 3-D tensor derivatives, computing 4-D tensor derivatives is generally untenable.
    Instead, numerical methods like finite differences for gradients and BFGS \cite{bfgs_1970} for Hessians can perform at reasonable speed.
    In order to approach real-time computation, it is likely that the full Hessian of $\chi$ is not necessary to find an appropriate search direction and that a partial computation or even leaving out the Hessian completely is sufficient to compute optimal trajectories.
    Future work will address this gap.
    Nonetheless, the algorithm in its current form can still be useful for offline planning for trajectories that are expected to have a high degree of risk, such as leaping across ledges or traversing narrow beams.
    In real-world applications, it can be acceptable for a robot to pause and plan a safe trajectory before executing these dangerous maneuvers.

\begin{table*}[t]
    \centering
    \caption{Rocket Hopper Convergence Measure and Simulation Results}
    \label{table:slip_results}
    \setlength{\tabcolsep}{4pt}
    \begin{tabular}{c|cccc|ccc|ccc|ccc}
        \multirow{2}{*}{Trial} &
        \multicolumn{4}{c|}{LQR Parameters}   & 
        \multicolumn{3}{c|}{Convergence Measure} & 
        \multicolumn{3}{c|}{Mean Simulated Error Ratio} & 
        \multicolumn{3}{c}{Mean Simulated Feedback Effort}\\ 
        \cline{2-14}
        & $Q_{\chi}$ & $Q_N$ &  $R_{i_{\text{air}}}$ & $R_{i_{\text{stance}}}$
        & Vanilla & $\chi$-iLQR & \%Difference & Vanilla & $\chi$-iLQR & \%Difference & Vanilla & $\chi$-iLQR & \%Difference \\
        \thickhline
        1 & 50 & 500$I$ & $0.01I$ & $0.1I$ & 1.01 & 0.71 & -29.70\% & 0.42 & 0.33 & -21.72\% &$1.74\cdot10^{-5}$ & $1.6\cdot10^{-5}$ & -7.16\%\\

        2 & 50 & 800$I$ &  $0.005I$ & $0.01I$ & 0.78 & 0.51 & -34.50\% & 0.32 & 0.24 & -25.74\% &$4.66\cdot10^{-5}$ & $3.25\cdot10^{-5}$ & -30.31\%\\
        
        3 & 50 & 250$I$ & $0.02I$ & $0.05I$& 1.14 & 0.94 & -17.70\% & 0.49 & 0.45 & -8.00\% &$7.12\cdot10^{-6}$ & $7.05\cdot10^{-6}$ & -1.01\%\\

        4 & 75 & 500$I$ &  $0.01I$ & $0.01I$ & 1.01 & 0.68 & -33.24\% & 0.41 & 0.32 & -21.73\% &$1.89\cdot10^{-5}$ & $1.62\cdot10^{-5}$ & -14.20\%\\
    \end{tabular}
\end{table*}

\section{Examples and Results}

In this section, we demonstrate the convergence improvements of our method on a spring hopper system and a planar quadruped robot model.
Simulation results show that the improved convergence measure correlates with an improvement in average tracking performance, robustness to large disturbances, and feedback control effort.
Both examples were implemented in MATLAB, with forward simulations using the \verb|ode113| function.
Cost function gradients were computed using \eqref{eq:product_rule}, derivatives of $\tilde{A}_i$ and $\Xi_{(i,i+1)}$ were computed with finite differences, and Hessians were computed with BFGS.

\subsection{Rocket Hopper}

\subsubsection{Rocket Hopper Model}
This system is made up of a point mass body with a single massless spring leg.
The state of the hopper is characterized by the positions $x_B$, $y_B$ of the body, the angle $\theta$ of the leg and their derivatives $\dot{x}_B$, $\dot{y}_B$, $\dot{\theta}$ such that the full state is a $6\times 1$ vector.
The system has two domains: an aerial phase $D_1$ and a stance phase $D_2$.
Taking a constant ground height at zero gives a touchdown guard function $g_{(1,2)}$ that is the height of the foot and a liftoff guard function $g_{(2,1)}$ that is the ground reaction force applied by the spring leg.
Both reset maps $R_{(1,2)}$ and $R_{(2,1)}$ are identity since position and velocity are continuous.

The system has two inputs: a hip actuation and an actuation in the direction of the leg.
In the air, this allows the hopper to rotate the leg around the body and exert a propulsion in the direction of the leg, somewhat akin to a rocket, though this force can approximate forces from other legs or actuators.
A small rotor inertia in the air ensures the dynamics are well-conditioned when controlling the massless leg.
In stance, the hip torque and rocket force exert ground reaction forces on the body.
The body mass of the hopper was chosen as 1 kg, spring constant as 250 N/m, and resting leg length as 0.75 m.

\subsubsection{Rocket Hopper Results}

The objective for this system is to begin in the air at rest with a height of 2 m and end in the air at rest with the same height displaced 0.2 m horizontally.
The system is given 1.5 s for this trajectory.

We generated four trials of paired trajectories with varied weighting parameters, shown in Table \ref{table:slip_results}, and compared the performance of the standard (vanilla) iLQR method (where $Q_{\chi} = 0$) to $\chi$-iLQR.
There is no reference trajectory to track, so $Q_i$ is zero for all trials.

For 3 of these trials, vanilla iLQR generated a trajectory with $\chi > 1$, meaning the worst-case error direction was expansive, see Table \ref{table:slip_results}.
$\chi$-iLQR decreases every convergence measure to below 1 so that all error directions are reduced over the trajectory.
On average, $\chi$-iLQR decreased $\chi$ by 28.79\% compared to the vanilla method.

To validate these trajectories, each closed-loop trajectory was simulated 100 times with equivalent small random initial perturbations in both positions and velocities with covariance matrix $\text{cov}(X_0) = 10^{-4} I$.
A small covariance was chosen so that the linearizations assumed in the convergence measure and LQR control are valid.
For each simulation run, the initial error $\delta x_0$ and the final error $\delta x_f$ were recorded, along with the sequence of control inputs $V := \{v_0,v_1,...,v_{N-1}\}$.
Note that these inputs are distinct from the nominal feedforward inputs to the system $U$ because there is additional feedback effort exerted by the actuators. 

Two values were recorded during each simulation run to characterize the convergence properties of the trajectories.
The first is the error ratio, $E = \frac{||\delta x_f||_2}{||\delta x_0||_2}$ defined as the ratio of the final error 2-norm to the initial error 2-norm.
A lower error ratio means better tracking performance, and $E<1$ indicates a net reduction in error on average.
The second value is the feedback effort, $F = \sum_{i=0}^{N-1}{(v_i - u_i)^2}$ which is the sum of squares of the difference between $V$ and $U$.

Table \ref{table:slip_results} shows that the simulation results support our assertion that an improved convergence measure correlates with an improved mean tracking performance and feedback effort.
The mean error ratio and feedback effort over the 100 simulations were both lower for trajectories generated with $\chi$-iLQR.
The average improvement over the four trials was 19.30\% for mean error ratio and 13.17\% for feedback effort.

None of the simulated runs had an error ratio greater than one, which is sensible since the worst-case direction occurs with probability zero.
However, even if none of the sampled initial errors aligned exactly with the worst-case direction predicted by the fundamental solution matrix, nearby initial error directions still see improvement in convergence, which explains the improvement in mean simulated error ratio.

\subsection{Planar Quadruped}

Here we demonstrate the improvements of $\chi$-iLQR on a more complex robot model akin a standard quadruped robot.
The model is simplified as a planar quadruped, meaning that all movement occurs in the sagittal plane and the left-right pairs of legs are constrained to move identically.

\subsubsection{Planar Quadruped Model}

In the sagittal plane, we can model the robot with 7 positional states.
$x_B$, $y_B$, $\theta_B$ are the position and orientation of the body. The front and back sets of legs each have two states for the hip angle $\alpha_f, \alpha_b$ and knee angle $\beta_f, \beta_b$.
Thus the full state is dimension 14.

This system has four domains: the aerial domain $D_1$, front stance domain $D_2$, back stance domain $D_3$, and full stance domain $D_4$.
The impact guard function is the height of the foot and the guard function for liftoff is the vertical ground reaction force.
The dynamics of the robot body in the aerial phase follow ballistic motion, while the legs are simplified to be massless while including the aforementioned rotor inertia.
The impact reset map for each foot consists of a discrete update to the hip and knee velocities, while the body states are unchanged due to the massless legs. 
The liftoff reset map is identity.
The input vector for this system is 4-dimensional to actuate the hip and knee joints.

In this model, parallel torsion springs are added to the knee joints.
Parallel joint springs have been utilized to mimic tendons found in animals \cite{endo_muscle_tendon_2014} that increase the energy efficiency of legged locomotion 
\cite{scheint_parallel_spring_2010, mombaur_2009}.
Due to the resonance of the natural spring dynamics, controlling these systems requires special care \cite{lakatos_compliance_2018,ackerman_elastic_load_2013}.
For example, \cite{pollayil_soft_locomotion} solved for optimal gait timings to leverage resonant spring frequencies.
These spring models of legged robots are good candidates for $\chi$-iLQR because the dynamics of the stance phase depend strongly on the leg configuration at touchdown.
Thus, a small error in leg states at touchdown can have a large effect on tracking performance.

The inertial and dimensional properties were chosen to match the Ghost Robotics Spirit 40 quadruped.
The added torsional knee spring has a spring constant $75 \text{ N}\cdot\text{m}\cdot\text{rad}^{-1}$ and rest angle 1.2 rad.

\subsubsection{Planar Quadruped Results}

The trajectory optimization task for the planar quadruped is to generate a gait with a forward velocity of 0.25 m/s.
The robot begins in the air with a body height of 0.3 m.
The hip joints begin at an angle of 0.6 rad and the knee joints begin at 1.2 rad.
The terminal target state is translated 0.0875 m in the x-direction from the initial state.
The trajectory is given 0.35 s to execute.
We choose to set a constant input weight of $R_i = 5\cdot10^{-4} I$.
The terminal weight is $Q_N = 500 I$ and the convergence weight for $\chi$-iLQR is $Q_{\chi} = 1$.

We set up a similar experiment to the prior example, with the addition of simulating over a range of covariance magnitudes.
This is done to evaluate the basin of attraction of each trajectory over larger initial errors that introduce greater nonlinear effects.
The two trajectories were evaluated with 6 sets of 100 paired simulation runs with random initial error covariance magnitudes of $10^{-4}$, $5\cdot10^{-4}$, $10^{-3}$, $5\cdot10^{-3}$, $10^{-2}$, and $5\cdot10^{-2}$ in each direction.
The lowest covariance magnitude of $10^{-4}$ approximates local linear behavior well, while $5\cdot10^{-2}$ is the maximum magnitude before some trials begin with the robot's feet below the ground.

Table \ref{table:quad_results} shows the vanilla cost \eqref{eq:vanilla_cost}, convergence measure, mean simulated error ratio, and mean simulated feedback effort of the two trajectories at the covariance magnitude $10^{-4}$.
As expected, the vanilla cost of the convergent trajectory increases since its optimizing for a different cost function, while the convergence measure and simulation values improve.
We argue that in dynamic legged locomotion, a costlier nominal trajectory can often be worth an improvement in the trajectory's robustness.

\begin{figure}[t]
    \centering
    \includegraphics[width=1\linewidth]{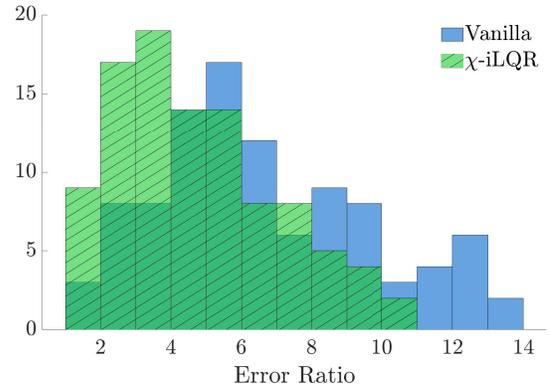}
    \vspace{-2em}
    \caption{Histogram of error ratio for 100 paired simulated trials of the quadruped model with small initial perturbations. Error ratio is the 2-norm of final errors divided by the 2-norm of initial errors.}
    \label{fig:quad_error_ratio}
\end{figure}

\begin{table}[t]
     
    \caption{Mean convergence results for the quadruped model for vanilla and $\chi$-iLQR with covariance magnitude $10^{-4}$}
    \label{table:quad_results}
    \centering
    \begin{tabular}{ cccc } 
    
      & Vanilla & $\chi$-iLQR & \%Difference\\
     \thickhline
     Vanilla Cost & 65.58 & 74.39 & +13.43\%\\ 
     \hline
     Convergence Measure & 60.52 & 41.35 & -31.68\% \\
     \hline
     Mean Simulated Error Ratio & 6.66 & 4.78 & -28.23\% \\
     \hline
     Mean Simulated Feedback Effort & 0.016 & 0.013 & -16.56\% \\
     \hline
    \end{tabular}
    
\end{table}
The mean simulated error ratio for the convergent trajectory at this small covariance magnitude was 28.23\% less and the mean simulated feedback effort was 16.56\% less.
Fig. \ref{fig:quad_error_ratio} displays a histogram of the error ratio for each of the trials, with the convergent trajectory having improved performance.

\begin{figure}[t]
    \centering
    \includegraphics[width=1\linewidth]{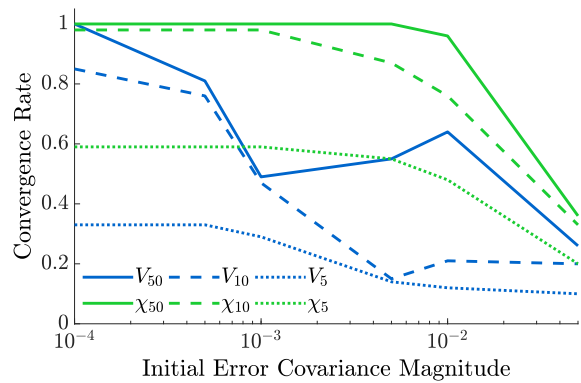}
    \vspace{-2em}
    \caption{Plots showing success of LQR controllers at tracking vanilla ($V$) and convergent ($\chi$) trajectories over various initial perturbation covariances. $V_{50}$ and $\chi_{50}$ indicate proportion of trials where error ratio was below 50, $V_{10}$ and $\chi_{10}$ below 10, and $V_{5}$ and $\chi_{5}$ below 5.}
    \label{fig:quad_convergences}
\end{figure}

As the magnitude of initial errors grows, the performance of the LQR tracking controller becomes worse due to the increase in nonlinear effects. 
Fig. \ref{fig:quad_convergences} shows the simulation results for each trajectory over a range of initial error covariance magnitudes.
Each pair of lines indicates the success rate of the respective closed-loop trajectories at maintaining error ratios of less than 50, 10, and 5 respectively.
An error ratio of greater than 50 is representative of a catastrophic failure, which the vanilla trajectory encounters at a covariance magnitude of $5\cdot10^{-4}$, while the convergent trajectory first experiences a failure at covariance magnitude $10^{-2}$.
This difference in performance suggests the convergent trajectory is more robust to larger initial errors and nonlinearities.

Even with the $\chi$-iLQR convergence improvements, the controller does not reduce errors in all directions.
The simulation results show there was usually some error growth, which is reasonable since the body dynamics are fully unactuated in the aerial phase and the system undergoes multiple hybrid events.
A combination of higher feedback gains and a global footstep planner could be able to grant this system full convergence.
Even so, this work can be valuable to ensure that the system does not diverge too far from its target trajectory between iterations of a global planner.

\addtolength{\textheight}{-4cm}

\section{Conclusion}
In this work, we present a novel trajectory optimization method, $\chi$-iLQR that optimizes over the worst-case error growth of a hybrid trajectory.
This method is based on the fundamental solution matrix, which maps the evolution of perturbations through a trajectory.
Incorporating the saltation matrix into the fundamental solution matrix allows for straightforward handling of hybrid events.
The simulation results presented on two legged robot models demonstrate that this method produces trajectories with improved tracking performance, decreased feedback actuation effort, and improved robustness to large perturbations.
Even for a quadrupedal trajectory that was very difficult to track, $\chi$-iLQR produced a trajectory that was superior at avoiding failures.

Following this work, we aim to apply these principles to robot hardware and demonstrate robust performance for maneuvers such as leaping and flipping.
We also aim to apply this work to other hybrid systems like robots that undergo stick-slip transitions.
This work and its extensions will further enable robots to navigate complex, uncertain environments and unlock worlds for robots to explore.

\bibliographystyle{IEEEtran}
\bibliography{ref}

\end{document}